\documentclass[pmlr]{jmlr}

\usepackage{longtable}

\usepackage{booktabs}
\usepackage[load-configurations=version-1]{siunitx} 

\usepackage{dirtytalk}

\jmlrvolume{-}
\jmlryear{-}
\jmlrworkshop{-}

\title[Playing Minecraft with Behavioural Cloning]{Playing Minecraft with Behavioural Cloning}

  \author{
   \Name{Anssi Kanervisto} \Email{anssk@uef.fi}\\
     \addr School of Computing, University of Eastern Finland\\
     \addr Joensuu, Finland
   \AND
   \Name{Janne Karttunen} \Email{janne.a.karttunen@gmail.com}\\
     \addr School of Computing, University of Eastern Finland\\
     \addr Joensuu, Finland.
     \AND
     \addr Karelics \\
     \addr Joensuu, Finland
   \AND
   \Name{Ville Hautam\"aki} \Email{villeh@uef.fi}\\
     \addr School of Computing, University of Eastern Finland\\
     \addr Joensuu, Finland
}

\begin{document}

\maketitle

\begin{abstract}
MineRL 2019 competition challenged participants to train sample-efficient agents to play Minecraft, by using a dataset of human gameplay and a limit number of steps the environment. We approached this task with behavioural cloning by predicting what actions human players would take, and reached fifth place in the final ranking. Despite being a simple algorithm, we observed the performance of such an approach can vary significantly, based on when the training is stopped. In this paper, we detail our submission to the competition, run further experiments to study how performance varied over training and study how different engineering decisions affected these results.
\end{abstract}


\section{Introduction}
    \textit{Reinforcement learning} (RL) is notorious for being sample inefficient and for providing different results on different training runs \cite{henderson2018deep}, but it can be supported with \textit{imitation learning}, like \textit{behavioural cloning} (BC) \cite{pomerleau1989alvinn, bojarski2016end, vinyals2019grandmaster, de2019causal}, to kickstart the learning process and reduce number of training samples needed. To support such research, MineRL 2019 competition \cite{mineRLcompetition} challenged participants to train agents to play Minecraft with limited amount of training time in the environment, along with a dataset of human gameplay to learn from \cite{guss2019minerldata}.
    
    Minecraft is an open-world, 3D vision-based video game where players progress by collecting resources and crafting tools, which enables harvesting of further resources. The world is procedurally generated at the start of each game, creating a new experience for each game. This inherent randomness, vision-based gameplay, hierarchical progression and open-ended nature makes Minecraft a good test-bench for new RL and imitation learning methods \cite{johnson2016malmo, guss2019minerldata}. MineRL competition challenges players to obtain diamond in the game, a feat that takes an experienced human player $5$ to $15$ minutes to complete \cite{guss2019minerldata}. On top of allowing only a limited training budget, submission systems were \textit{trained} on the evaluation server, encouraging the use of \textit{robust} methods that provide the same results from different training runs.
    
    Motivated by these limitations and recent results in human-level BC in Starcraft II \cite{vinyals2019grandmaster}, we began our work on the challenge by only using behavioural cloning, \textit{i.e.} by predicting what actions human players would take. We believed BC would be a robust alternative to RL methods, and also perform well enough to be competitive. But we learned that BC requires the same level of engineering for stability and performance, despite being a simple method. While this is not a novel observation \cite{de2019causal}, BC results are often represented as single, averaged numbers without further detail on variance, painting a picture of a stable learning method \cite{hester2018deep, vinyals2019grandmaster, bojarski2016end, codevilla2018end}.
    
    In this paper, we summarize our submission to the MineRL 2019 competition, based on BC, and study the issues we ran into during the competition. Specifically, we discuss the variance in agent's performance during training period, the effect of uniform sampling of the training dataset, the use of data augmentation to improve the performance and the possibility of agent biasing towards over-represented actions. Our main contribution is highlighting how behavioural cloning is not as robust as expected, and how we should also report variance in its results, just like in RL research \cite{henderson2018deep}.

\section{MineRL competition}
    Contrary to previous Minecraft-related competitions, like the MARL\"O competition \cite{perez2019multi}, MineRL competition's task requires player to complete a hierarchical \textit{crafting-tree} by harvesting resources and crafting items \cite{mineRLcompetition}. Agents were rewarded with exponentially increasing rewards as they progressed in this crafting tree, and sum of rewards per game were used as an evaluation metric, averaged over $100$ games. In the first round (\say{Round 1}), participants had to submit their code along with a trained agent to the evaluation server where it was evaluated. In the second round (\say{Round 2}), ten finalists \textit{only submitted their program code} and the agent was trained on the evaluation server.
    
    The provided MineRL package consists of two distinct parts: A dataset of human plays in Minecraft \cite{guss2019minerldata}, and a corresponding RL learning environment built on top of Malmo \cite{johnson2016malmo}. Learning agents are provided with similar observations and actions a human player would have: a vision observation from the point-of-view of the character, information on the contents of current inventory and actions as keyboard-like on-off decisions, along with the horizontal and vertical turning of the camera. After each action, the game progresses by $1/20$ seconds. On top of the standard movement controls (\textit{e.g.} forward, backward, move camera, jump), MineRL exposes convenience functions for \textit{e.g.} crafting with a single-step action, while a human player has to perform a complex GUI to craft the desired item. The full action space of \texttt{MineRL-ObtainDiamond-v0} task consists of $13$ discrete variables with different number of options, as well as $2$ continuous actions for moving camera on both axes. The dataset provides all observations and actions in these same formats.

\section{Playing Minecraft with Behavioural Cloning}
    Upon seeing the evaluation protocol, we realized RL training could be too unreliable \cite{henderson2018deep}. Behavioural cloning, on the other hand, is easy to implement and has been applied to practical problems with success \cite{vinyals2019grandmaster, bojarski2016end}. With all its successes, it is also known to suffer from \textit{distributional shift} and \textit{causal confusion} \cite{de2019causal}, where model fails to learn the true causal-effect relationship between observations and actions. Former of these are commonly cited as the main limitation of behavioural cloning, as it breaks the i.i.d. assumption of supervised learning by subjecting the agent to different distributions of observations during training and testing \cite{ross2011reduction}. To combat this, methods like DAgger \cite{ross2011reduction} gradually gather expert demonstrations by playing the environment. Alas, such methods are infeasible without an access to an expert, and competition rules prohibited using additional data. Another approach would be to use \textit{batch reinforcement learning} \cite{fujimoto2019benchmarking}, where RL agents are trained with a fixed dataset, but the performance of such methods is unclear (see benchmarks of \cite{fujimoto2019benchmarking} versus the results in original articles per method). 
    
    With these observations in mind, we started our work on the competition submission using BC, with the goal to obtain similar human-like performance as in Starcraft II. Our final submission using only BC reached rank \#5/10 in the final round (from a total of $40$ participants in the whole competition). Overview of our system is illustrated in the Figure \ref{fig:system}. The round 1 submission did not include a replay buffer, the effect of which we discuss in Section \ref{sec:variance}. Code to this submission is available at \url{https://github.com/Miffyli/minecraft-bc}.

    \begin{figure}[tbp]
        \floatconts
        {fig:system}
        {\caption{Our submission to the MineRL competition, for playing Minecraft with behavioural cloning. Action heads are treated independent from each other.}}
        {\includegraphics[width=1.0\linewidth]{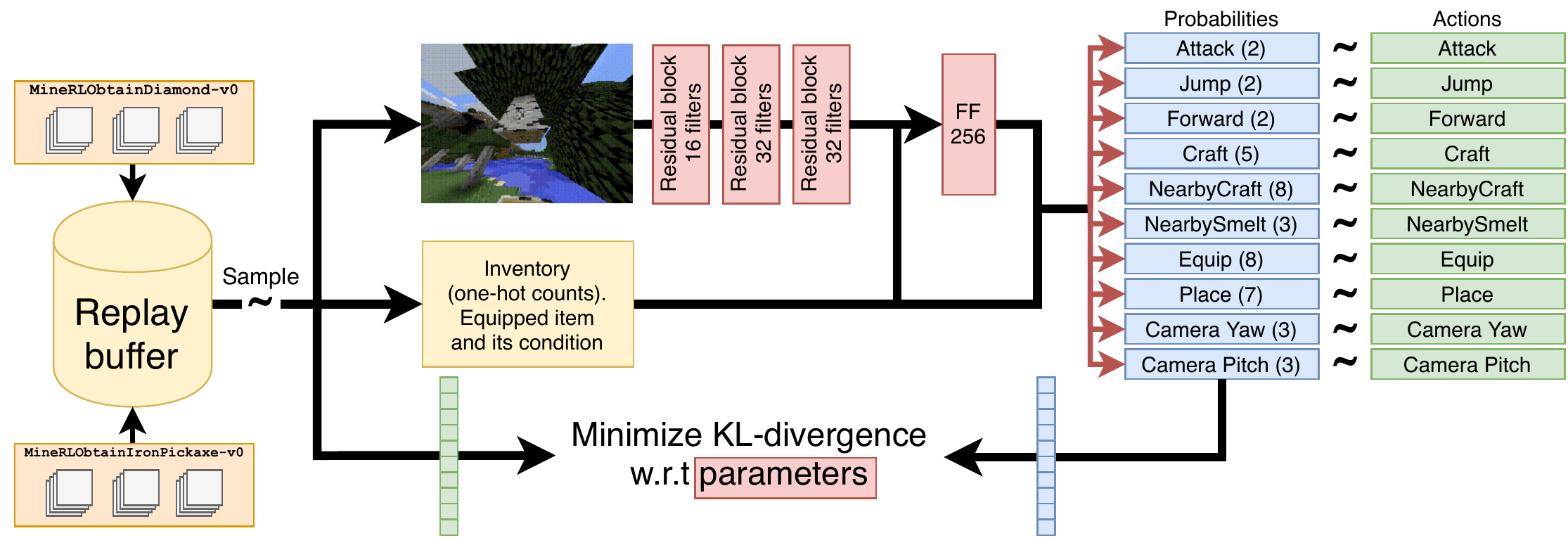}}
    \end{figure}

    \subsection{Observations}
        Observations consist of two parts: A visual observation and \say{direct features}. Visual observation, corresponding to what a human player would see on the screen, is a RGB image of resolution $64 \times 64$. An example of this is shown in Figure \ref{fig:system}. Direct features represent the amount of items in the inventory (one-hot vectors up to size of $8$), currently equipped item (one-hot vector) and durability of the equipped item (scalar in $[0, 1]$, zero being a broken item.

    \subsection{Actions}
        Each agent action consists of $10$ independent discrete decisions (referred as \say{actions}), each with varying number of options to choose from, corresponding to a \texttt{MultiDiscrete} space in OpenAI Gym \cite{brockman2016openai} and illustrated in Figure \ref{fig:system} on the right. We removed the following actions to simplify action-space: \texttt{move left}, \texttt{move right}, \texttt{move back}, \texttt{sneak} and \texttt{sprint}. Moving the mouse is discretized into three options per axis (turn left/up, turn right/down or stay still for that axis), with fixed movement speed of two degrees per step. Each action is repeated for four steps ($0.2s$ of gameplay). During the evaluation, we replace actions to craft axes (used to cut trees faster) and un-equip the main-hand item with \say{no-ops}, latter of which improved results as agent would not be able to accidentally un-equip crucial items for progression (pickaxes for mining).
        
        We process dataset samples by discarding the same actions as above by simply ignoring them. We convert the continuous mouse movement into discrete by thresholding: if a player moved the mouse by more than one degree to the right, then discrete mouse action for pitch is to move right, for example.
        
    \subsection{Model architecture}
        Agent's model for predicting actions is a deep neural network, following the architecture presented in \cite{espeholt2018impala}. Image observations are processed with a residual network, consisting of three residual blocks \cite{he2016identity}, concatenated with direct features, fed through a single fully-connected layer, concatenated with direct features again and then mapped into probabilities per action with softmax activation. Options for each action are then sampled from these probabilities (multilabel classification). All layers are followed by a ReLU activation and initialized with FixUp method \cite{zhang2019fixup}. Direct connection between direct features and action probabilities allows inventory counts to directly emphasize actions, \textit{e.g.} the player should always craft logs into planks.
        
        Note that we do not include any recurrent network techniques like LSTM or provide past frames, like done in \textit{e.g.} Atari games \cite{mnih2015human}. Related work with BC has shown providing such information to be detrimental to the performance \cite{wang2019monocular, de2019causal}. We observed the same in our internal experiments, both in this competition and during our participation in the Obstacle Tower Challenge \cite{juliani2019obstacle}. In addition, this prevents the confusion with dropped actions by the action-space, \textit{e.g.} \say{player did not press anything, yet they moved backward?}. Theoretically, none of the tasks in the challenge require memory to solve.

    \subsection{Training}
        The agent is trained on the \texttt{MineRLObtainDiamond-v0} and \texttt{MineRLObtaindIronPickaxe-v0} subsets of the dataset for $25$ epochs. We store samples into a replay buffer of size $500 000$, from which we sample batches used for training (further discussed in the Section \ref{sec:variance}). We discard all the samples where the player took no action. Parameters are updated using an Adam optimizer \cite{Adam} with a small learning rate of $5 \cdot 10^{-5}$ for more stable learning, and with L2 regularization weight of $10^{-5}$. Network is trained to minimize KL-divergence between predictions and actions from the dataset, given the single observation. One-hot action labels are smoothed with a smoothing constant $0.005$ to avoid overfitting \cite{szegedy2016rethinking}. We found this to improve performance. In total, the network is trained with approximately $2.3$ million batches of $32$ elements.  

        To improve generalization, we augment the dataset by applying random noise, random adjustments to brightness and contrast and by randomly flipping the images horizontally. We further discuss this in Section \ref{sec:augmentation}.

    \subsection{Competition results}
        
        \begin{figure}[tbp]
        \floatconts
          {fig:minerl_results}
          {\caption{Distribution of progress in the crafting tree obtained by our BC systems in rounds 1 and 2. Y-axis shows how often the agent reached the item on X-axis. Round 1 submission had higher average reward, but round 2 submission obtained cobblestone and stone pickaxes more reliably after wooden pickaxe.}}
          {%
            \subfigure[Round 1]{\label{fig:round1_results}%
              \includegraphics[width=0.45\linewidth]{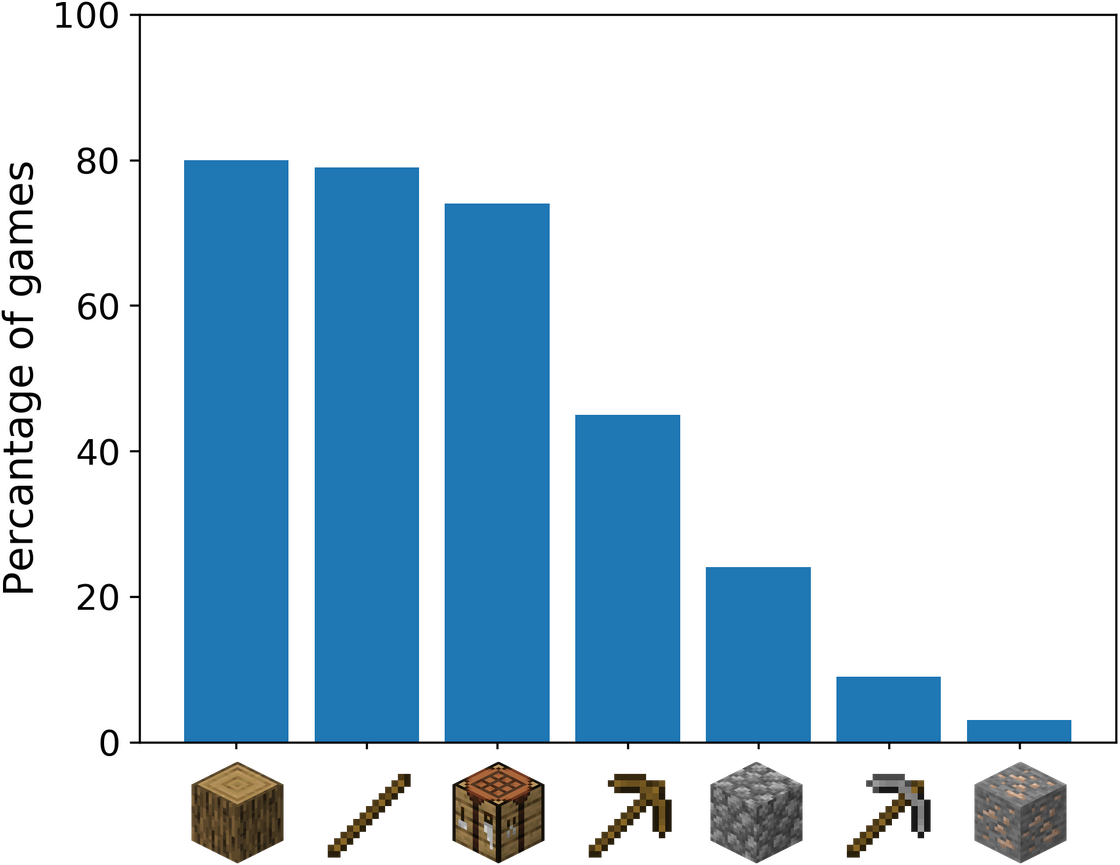}}%
            \qquad
            \subfigure[Round 2]{\label{fig:round2_results}%
              \includegraphics[width=0.45\linewidth]{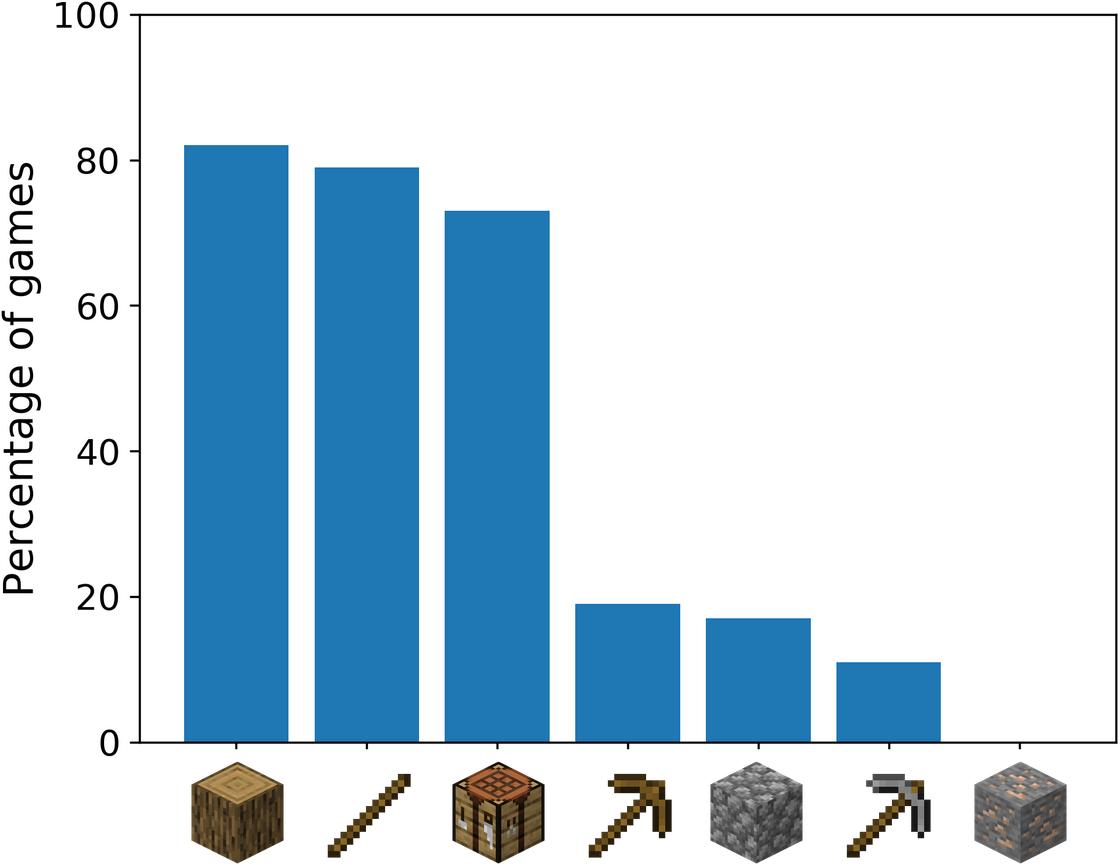}}
          }
        \end{figure}
    
        Our round 1 submission scored an average reward of $21.7$ and round 2 submission $17.9$. We found evaluation results to vary significantly between different evaluation runs, even with $100$ games used to evaluate the agents by the evaluation server.

        Figure \ref{fig:minerl_results} shows the distribution of how far our round 1 and 2 submission progressed in the crafting tree. In round 1, half of the games where agent obtained a crafting table also obtained a wooden pickaxe, while in round 2 only one-fifth achieved the same. After the wooden pickaxe, round 2 submission obtained cobblestone and a stone-pickaxe more reliably than the round 1 submission. This suggests the round 2 submission could have achieved higher reward, had it only learned to craft a wooden pickaxe.
        
        Upon discussing with the other participants, we learned that at least two teams with higher or equal scores also used BC without RL\footnote{Private communication.}. This leads to two conclusions: \textbf{1)} Behavioural cloning can be used to play Minecraft, with comparable performance to RL methods. \textbf{2)} Such performance requires careful engineering, as also summarized by \cite{de2019causal}.

\section{Discussion and further analysis}
    
    \subsection{Variance in the results}
        \label{sec:variance}
        
        As this behavioural cloning can be seen as a multilabel classification task, we did not expect the evaluation performance to vary from run-to-run, with a steady improvement in performance over training. After all, many published research articles with results with BC only show single, averaged numbers without variance (\textit{e.g.} \cite{hester2018deep, vinyals2019grandmaster, bojarski2016end, codevilla2018end}). Turns out this was not so.
        
        \begin{figure}[tbp]
            \floatconts
            {fig:learning_curve}
            {\caption{Learning curves of six different training runs with the round 2 submission code, three for shorter and three for longer training. Each evaluation point is an average over $200$ games (standard variance in $[10, 16]$ for agents after million updates). The first evaluation is after $25000$ updates. Solid line is mean over the three runs, with individual runs shown in transparent lines. The performance of an agent can be significantly lower or higher, depending on when the training is stopped.}}
            {\includegraphics[width=1.0\linewidth]{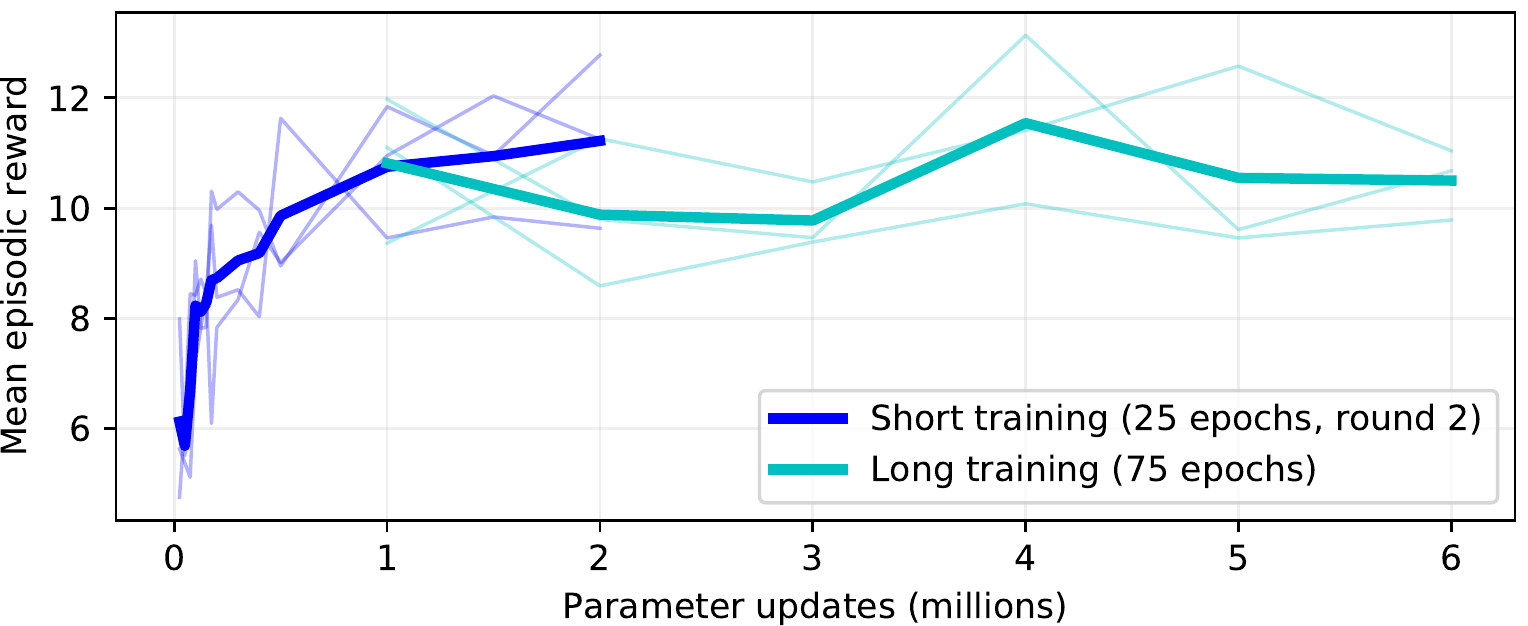}}
        \end{figure}
        
        Figure \ref{fig:learning_curve} shows learning curves of six individual runs, split into three shorter and three longer runs\footnote{Local evaluation performance is significantly lower than on competition evaluation servers for unknown reasons. Other competition participants have reported the same (private communication).}. We include longer runs to confirm the agent does not improve with longer training. Most differences between consecutive evaluation points are not statistically significant according to the t-test (two-tailed, $p > 0.05$)\footnote{We assume fixed sample mean variance, given the narrow range of means we focus on here. Larger sample means have larger variance, due to exponential spacing of individual rewards.}, as the variances of evaluations are large. There is also no consistent improvement after two million updates. That said, note the sudden spikes, especially in the beginning and at the four million updates. By stopping the training at the right time, we can jump from an average reward of $9.5$ to $13.1$, a relative increase of $38\%$ (statistically significant change with $p < 0.05$). The same was observed with learning rate annealing to zero. This demonstrates the need to also study variances in results with BC methods, as done in \textit{e.g.} \cite{de2019causal}.

        One explanation for the variance in our case is the non-uniform sampling of the dataset. The MineRL dataset consists of videos, one per game (or episode). The provided dataset-loader reads consecutive frames of trajectories, leading to correlated batches of training samples, despite reading samples concurrently from many games. We observed this as a periodic oscillation in the training loss throughout the training. To balance the sampling, we asynchronously load samples to a replay buffer of $500 000$ samples, from which we sample the training batches. With this setup we obtain an average loss of $1.505$ with std. $0.0281$ in the last $10\%$ of training updates (loss starts from $\approx 3.0$). By comparison, with a replay buffer of size $10 000$ (simulating the round 1 submission), we have a mean loss of $1.544$ and std. of $0.1136$, a four times higher variance. Doubling the buffer size did not reduce the variance. This shows that the original sampling had uneven training batches, and that a replay buffer helps to stabilize learning.

    \subsection{Augmenting dataset with noise}
        \label{sec:augmentation}
        Augmenting dataset by modifying images with noise, cropping, translation and other transformations is a very common practice in machine vision, RL and BC experiments alike (\textit{e.g.} \cite{codevilla2018end, karttunen2019video}). This simple trick improves the generalization of the trained agents, and it has been used in the previous RL video-game competitions (\textit{e.g.} \cite{dosovitskiy2016learning, unixpickle2019otc}). As such, we started our work by augmenting the dataset with the following transformations: Multiply pixels by uniformly sampled strength from $[0.9, 1.1]$ (contrast), with one value per channel to change the hue, add a uniformly sampled value from $[-0.1, 0.1]$ to all pixels (brightness), add normal noise to each pixel from $\mathcal N(0.0, 0.02)$, flip image (and associated actions) with $50\%$ probability and clip pixels to original range $[0.0, 1.0]$. The changes were visible to a human eye but did not obstruct any information. As such, we believed this would work for training.
        
        Turns out the augmentation was too strong. During competition we decreased the strengths to $[0.98, 1.02]$, $[-0.02, 0.02]$ and $\mathcal N(0.0, 0.005)$, respectively, which consistently improved the performance. With these strengths, the changes are hardly visible to a human eye, only visible in a side-by-side comparison. Repeating round 2 training with the original augmentations reached a score of $6.0$, a significantly lower result than all three round 2 results in the Figure \ref{fig:learning_curve}. After training with all of the possible combinations, all but one stayed below a score of $8$. The experiment with only brightness-augmentation obtained a score of $11.5$. This demonstrates that, haphazardly applying augmentation may not provide improved results, or even be detrimental, even if to a human eye the changes do not seem significant. 
    
    \subsection{Imbalanced dataset}
        \label{sec:bias}
        
        \begin{table}[tpb]
        \floatconts
          {table:class_inbalance}%
          {\caption{Distribution of options per action in \texttt{MineRLObtainDiamond-v0} subset, in the action space used by our system. First column corresponds to \texttt{None} action. Options in \texttt{craft}, \texttt{nearbyCraft} and \texttt{place} have very little representation, yet are crucial for progressing in the game (\textit{e.g.} ``craft a wooden pickaxe", ``place a crafting table").}}%
          {
            \begin{tabular}{l|llllllll}
                \textbf{Action} & \multicolumn{8}{c}{\textbf{Distribution of options per action (\%)}} \\
                \hline
                attack      & $49.16$ & $50.84$ &        &        &        &        &        &  \\
                camera\_x   & $84.77$ & $7.55$  & $7.68$ &        &        &        &        &  \\
                camera\_y   & $86.48$ & $6.65$  & $6.87$ &        &        &        &        &  \\
                craft       & $99.93$ & $0.01$  & $0.02$ & $0.03$ & $0.01$ &        &        &  \\
                equip       & $99.92$ & $\scriptsize{<}.01$  & $0.01$ & $0.02$ & $0.01$ & $0.03$ & $\scriptsize{<}.01$ & $0.02$                           \\
                forward     & $81.09$ & $18.91$ &        &        &        &        &        &  \\
                jump        & $96.54$ & $3.46$  &        &        &        &        &        &  \\
                nearbyCraft & $99.96$ & $\scriptsize{<}.01$  & $0.01$ & $\scriptsize{<}.01$ & $0.01$ & $\scriptsize{<}.01$ & $0.01$ & $0.01$                           \\
                nearbySmelt & $99.99$ & $0.01$  & $\scriptsize{<}.01$ &        &        &        &        &                                  \\
                place       & $99.52$ & $0.03$  & $0.06$ & $0.21$ & $0.02$ & $0.02$ & $0.14$ &                                                  
            \end{tabular}
        }
        \end{table}
        
        A \textit{class imbalanced} dataset can lead the classifier to bias towards classes that are over-represented \cite{chawla2002smote}. Same applies to BC, being a classification problem. Table \ref{table:class_inbalance} shows the ratio of actions (classes) in the  \texttt{MineRLObtainDiamond-v0} subset of the MineRL dataset. Special actions, like crafting, are barely represented, as these happen rarely in a single game: player only needs to craft a single crafting table, for example. Same applies even after removing all \textit{no-op} actions. We believed this would lead to the classical classification bias: network will learn to mainly predict the over-represented actions. This was the main motivation to include label smoothing.
        
        To study if class imbalance was an issue in our submission, we play $20$ games with the three short-train agents and record the taken actions. We discard samples where the action is not feasible, \textit{i.e.} when the player does not have the necessary materials to craft/place the item. To our surprise, some under-represented actions like crafting a wooden pickaxe ($0.01\%$ in the dataset) has constantly over $50\%$ probability of being sampled. However, crafting and placing a crafting table both have $99\%$ of the probabilities at under $60\%$, never going above $80\%$. Subjective analysis of these games show how the agent was able to reliably gather wood and craft it into a crafting table, but rarely places it down even when it needs it to craft further items. By supporting the agent with hard-coded actions for crafting, the evaluation results of the three short-train agents increased by relative of $30\%$, $37\%$ and $92\%$, all of which were statistically significant. Furthermore, the latter agent was able to reach a reward of $291$ in three of the $200$ games. This demonstrates how the agents were hindered by these specific actions, and again highlights the difference in agents, despite being trained with the same setup.
        
        During the competition we experimented with weighting the losses according to the rarity of labels in the dataset (rarer actions had higher weight), and with under-sampling (discard samples with only common actions). Neither of these improved the results. We believe a more sophisticated sampling technique, like SMOTE \cite{chawla2002smote}, could improve the results.

\section{Conclusion}
    We presented a behavioural cloning system for playing Minecraft, which reached fifth place in the final ranking, out of total of $40$ participants. This demonstrated its effectiveness among reinforcement learning methods, but during the competition we learned even BC is not free from the engineering required. 
    
    Digging deeper into the results, we learned that by stopping the training at the correct moment, the agent reaches a statistically significant improvement in the evaluation performance. We discussed the non-uniform sampling of the dataset as an explanation for this, and used a replay-memory to stabilize this. Meanwhile, agents had trouble executing actions that were under-represented in the dataset, leading to a lower performance. That being said, this was not observed for all such actions. All in all, we argue the research with BC should include variance of the results over multiple runs, just like with RL experiments \cite{henderson2018deep}. Work in \cite{de2019causal} is a great example of this.
    
    With many questions remaining, we believe there is still much untapped potential in BC. The better results of the two other contestants with BC are both uplifting and discouraging: we now know we could have done better, but at least we know we \textit{can} do better! This motivates us to study behavioural cloning further, both in regard to points brought up in this work, and in studying what was different in other competitors' submissions.

\bibliography{main}

\end{document}